\pdfoutput=1
\documentclass[conference]{IEEEtran}
\IEEEoverridecommandlockouts
\usepackage{cite}
\usepackage{amsmath,amssymb,amsfonts}
\usepackage{algorithmic}
\usepackage{graphicx}
\usepackage{textcomp}
\usepackage{xcolor}
\def\BibTeX{{\rm B\kern-.05em{\sc i\kern-.025em b}\kern-.08em
    T\kern-.1667em\lower.7ex\hbox{E}\kern-.125emX}}
\usepackage{generic}
\usepackage{algorithm}
\usepackage{hyperref}
\usepackage{multirow}
\usepackage{booktabs}
\usepackage{float}
\usepackage{placeins}
\usepackage{textcomp}
\usepackage{bm} 
\newcommand{\vA}{\bm{A}}
\newcommand{\vB}{\bm{B}}
\newcommand{\vC}{\bm{C}}
\newcommand{\vI}{\bm{I}}
\newcommand{\vx}{\bm{x}}
\newcommand{\vK}{\bm{K}}
\newcommand{\vy}{\bm{y}}

\begin{document}

\title{P-Mamba: Marrying Perona Malik Diffusion with Mamba for Efficient Pediatric Echocardiographic Left Ventricular Segmentation}

\author{
\IEEEauthorblockN{1\textsuperscript{st} Zi Ye}
\IEEEauthorblockA{\textit{Institute of Intelligent Software} \\
\textit{Guangzhou, China}\\
yezi1022@gmail.com}
\and
\IEEEauthorblockN{2\textsuperscript{nd} Tianxiang Chen}
\IEEEauthorblockA{\textit{University of Science and Technology of China} \\
\textit{China}\\
txchen@mail.ustc.edu.cn}
\and
\IEEEauthorblockN{3\textsuperscript{rd} Fangyijie Wang}
\IEEEauthorblockA{\textit{School of Medicine} \\
\textit{University College Dublin}\\
Dublin, Ireland \\
fangyijie.wang@ucdconnect.ie}
\and
\IEEEauthorblockN{4\textsuperscript{th} Hanwei Zhang}
\IEEEauthorblockA{\textit{Saarland University} \\
\textit{Germany}\\
zhang@depend.uni-saarland.de}
\and
\IEEEauthorblockN{5\textsuperscript{th} Lijun Zhang}
\IEEEauthorblockA{\textit{SKLCS, Institute of Software} \\
\textit{University of Chinese Academy of Sciences}\\
China \\
zhanglj@ios.ac.cn}
}

\maketitle

\begin{abstract}
In pediatric cardiology, the accurate and immediate assessment of cardiac function through echocardiography is crucial since it can determine whether urgent intervention is required in many emergencies. However, echocardiography is characterized by ambiguity and heavy background noise interference, causing more difficulty in accurate segmentation. Present methods lack efficiency and are prone to mistakenly segmenting some background noise areas, such as the left ventricular area, due to noise disturbance. To address these issues, we introduce P-Mamba, which integrates the Mixture of Experts (MoE) concept for efficient pediatric echocardiographic left ventricular segmentation. Specifically, we utilize the recently proposed ViM layers from the vision mamba to enhance our model's computational and memory efficiency while modeling global dependencies. In the DWT-based Perona-Malik Diffusion (PMD) Block, we devise a PMD Block for noise suppression while preserving the left ventricle's local shape cues. Consequently, our proposed P-Mamba innovatively combines the PMD ’s noise suppression and local feature extraction capabilities with Mamba’s efficient design for global dependency modeling. We conducted segmentation experiments on two pediatric ultrasound datasets and a general ultrasound dataset, namely Echonet-dynamic, and achieved state-of-the-art (SOTA) results. Leveraging the strengths of the P-Mamba block, our model demonstrates superior accuracy and efficiency compared to established models, including vision transformers with quadratic and linear computational complexity.
\end{abstract}


\section{Introduction}
\label{sec:introduction}
\IEEEPARstart{C}{ongenital} heart diseases (CHD) pose significant health risks, necessitating precise diagnostic tools like the Echocardiogram for early detection and treatment in children \cite{ref_vanderlinde2011}. Among these indices, the Left Ventricular Ejection Fraction (LVEF) is the most commonly used and vital metric for assessing systolic function \cite{b3}. The LVEF is predominantly calculated using the biplane Simpson’s standard protocol method in clinical settings\cite{b4}. This technique involves the manual delineation of the left ventricular endocardium by physicians in specific frames of the apical two-chamber (A2C) and apical four-chamber (A4C) echocardiographic views, a process essential for the identification of the end-systolic volume (LVESV) and end-diastolic volume (LVEDV). 

\begin{figure}[!t]
\centerline{\includegraphics[width=\columnwidth]{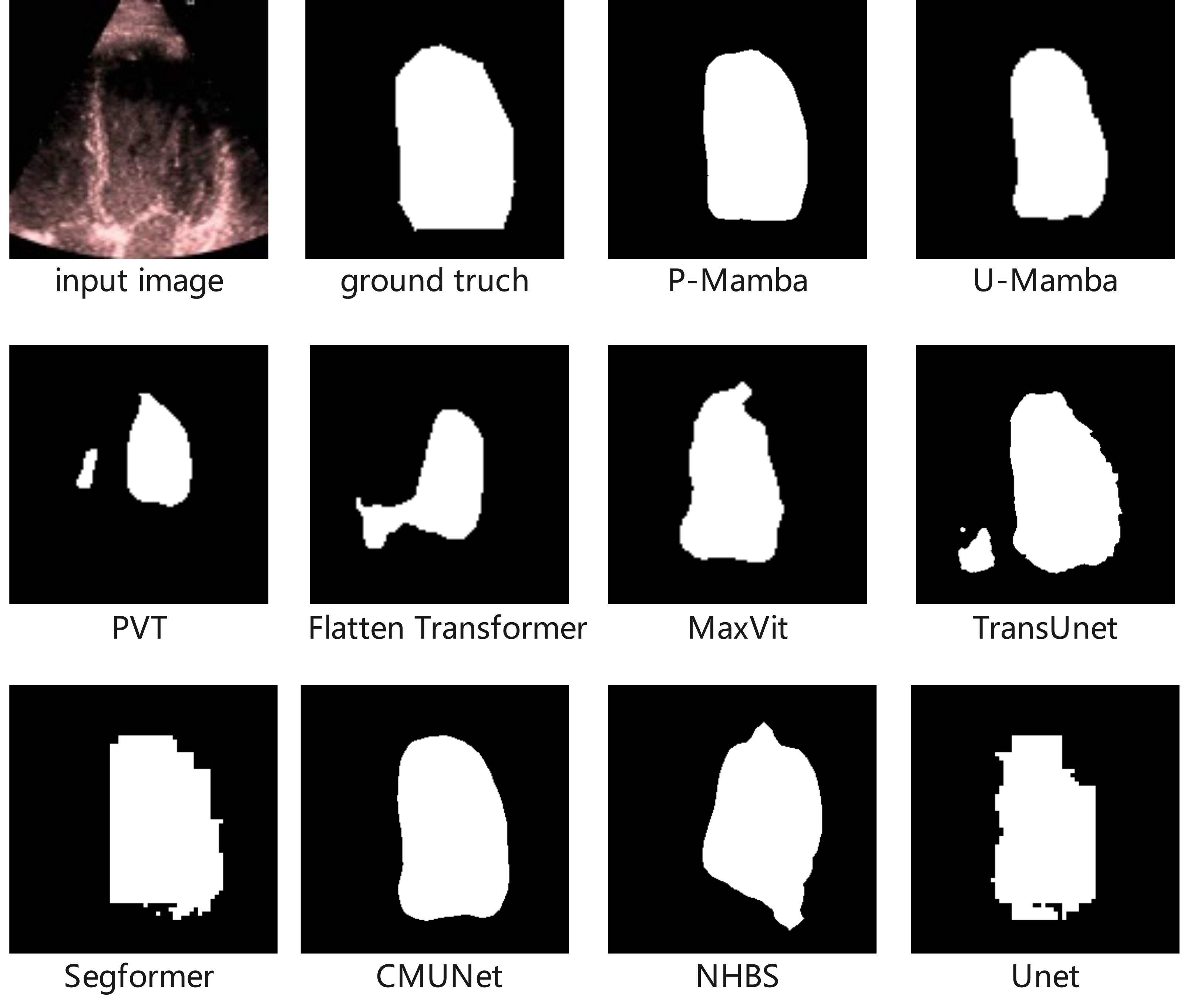}}
\caption{Visualization of the noise interference challenge among P-Mamba and other models on the pediatric A4C dataset.
\label{fig:noise}}
\end{figure}

Machine learning and artificial intelligence have significantly improved the dependability and precision of evaluating left ventricular (LV) function using echocardiography in adults, as shown by multiple research projects. However, machine learning is more difficult in youngsters due to diverse anatomical anomalies, heart rate, stature, and cooperative capacity. Various factors influence the spatial and temporal resolutions, eventually influencing echocardiographic imaging quality\cite{b5}. As a result, there is concern about how well machine learning models built on adult datasets can be applied in pediatric echocardiography owing to the more significant variabilities.

State Space Models (SSMs) have recently extracted broad interest from researchers. As the first proposed basic model built by SSMs, Mamba \cite{mamba} has achieved superior performance compared to transformers in long-range dependency modeling, even with linear complexity. Vision Mamba \cite{visionmamba} was later proposed to apply Mamba to the vision domain and achieves superb efficiency-accuracy balance compared with DeiT. U-Mamba \cite{umamba} was the first Mamba-based medical image segmentation method and boasted its efficient design. Based on the above works, we try to apply the Mamba structure to our model to guarantee model efficiency. 

\begin{figure}[h]
\centering
\includegraphics[width=\linewidth]{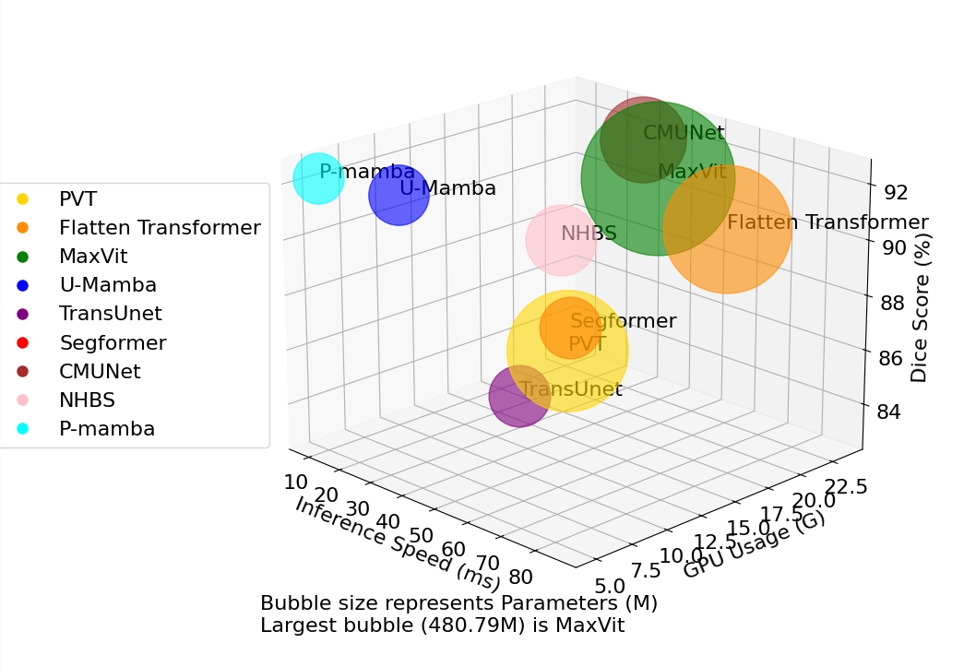}
\caption{The efficiency comparisons between P-Mamba and other models on the pediatric PSAX dataset}
\label{efficientcy}
\end{figure}

On the other hand, echocardiography often encounters challenges, including significant speckle noise, limiting its imaging technique. As shown in Fig.~\ref{fig:noise}, current approaches easily mistake some background noise areas for the target area. To suppress background noise while maintaining target structural features, we draw inspiration from Perona–Malik Diffusion (PMD) \cite{ref_perona1990}, initially in de-noising tasks to achieve this goal. Therefore, we tailor P-Mamba for more efficient and accurate pediatric echocardiographic LV segmentation. This model can eliminate noise while preserving the local target boundary details for the best performance. At the same time, our model demonstrates superior efficiency. 

Our main contributions are as follows:

\begin{itemize}
  \item We propose P-Mamba, which innovatively combines the PMD’s noise suppression and local feature extraction ability with the Vision Mamba’s efficient design for global dependency modeling and integrates the Mixture of Experts (MoE) concept to set new performance benchmarks on noisy pediatric echocardiogram datasets.
  \item Benefiting from PMD, our model excels by suppressing noise while preserving and enhancing target edges in ultrasound images, as shown in Fig.~\ref{fig:noise}.
  \item Extensive experiments demonstrate that our P-Mamba achieves superior segmentation accuracy and efficiency to other methods, including specialized ultrasound segmentation models \cite{cmu, nhbs}, vision transformers with quadratic \cite{ref_wang2021} and linear \cite{ref_han2023,ref_tu2022} computational complexity. In addition, Fig.~\ref{efficientcy} visually compares the efficiency among different models.

\end{itemize}

\section{Related Work}
\label{sec:Related Work}

To date, deep learning (DL) development has promoted automatic medical image segmentation, and several well-known deep learning frameworks have provided good ideas for echocardiography segmentation with outstanding performance.

\subsection{General Deep Learning Segmentation Methods}

Deep learning has revolutionized image segmentation in recent years, a crucial task in computer vision. Among the pioneering methods, U-Net \cite{ref_gupta2023} stands out for its effective encoder-decoder architecture, which captures context through its contracting path and refines localization through its expansive path. U-Net's simplicity and efficiency have made it a foundational medical image segmentation model.

Pyramid Vision Transformer (PVT) \cite{ref_wang2021} and TransUNet \cite{transunet} both draw significant inspiration from transformer architecture. PVT captures long-range dependencies through self-attention mechanisms and hierarchical feature representations, making it highly effective for dense prediction tasks. Similarly, TransUNet integrates transformer modules into the traditional U-Net structure, leveraging the transformer's attention mechanisms to enhance feature encoding and significantly improve segmentation accuracy, particularly in complex scenarios. Both models demonstrate the transformative impact of incorporating transformer principles into segmentation tasks.

UniFormer \cite{ref_li2023}, SpectFormer \cite{ref_patro2023}, and Segformer \cite{segformer} represent some of the most contemporary segmentation models. Segformer achieves state-of-the-art performance in multiple segmentation benchmarks with lower computing costs by integrating hierarchical transformers with effective feature fusion approaches. This simplifies the transformer architecture while maintaining its key benefits. UniFormer merges convolutional networks with transformers to balance local feature extraction and global context understanding, efficiently capturing multi-scale features to enhance segmentation performance. SpectFormer delves deeper into the transformer domain by incorporating spectral analysis, emphasizing frequency domain information to complement traditional spatial representations. These models exemplify the cutting-edge advancements in segmentation technology.

\subsection{Medical Image Segmentation Methods}

Recently, innovative deep-learning methods have become increasingly vital in medical analysis and represent the forefront of leveraging complex neural networks for precise medical image segmentation. For instance, CMU-Net \cite{cmu} and NHBS-Net \cite{nhbs} are specialized ultrasound segmentation models. CMU-Net uses hybrid convolution and multi-scale attention gates to improve feature extraction and context information. NHBS-Net has advanced attention and fusion modules, making clinical ultrasound applications more accurate and error-free. 

However, when applying the present segmentation methods to segment the left ventricle in pediatric echocardiograms, the challenge exists since the irregular shape of the left ventricles still cannot be well segmented since these methods pay insufficient attention to high-frequency boundary details, which can be seen in Fig.~\ref{fig:noise}. Also, present methods lack efficiency, hindering their wider application, as shown in Fig.~\ref{efficientcy}. Another concern of the studies above is the limitations of their reliance on proprietary datasets. The EchoNet-Peds dataset, developed by Stanford University, indicates the first publicly available pediatric echocardiography dataset \cite{b20}, featuring 4,467 echocardiograms from 1,958 patients, including a 43$\%$ female demographic and ages ranging from newborns to 18 years. This comprehensive dataset yielded 7,643 video clips and 17,600 labeled images, primarily from A4C and PSAX view clips. As a result, the video clips were strategically allocated, with 6,114 (80$\%$) for training, 765 (10$\%$) for testing, as well as 764 (10$\%$) for validation.

\subsection{Mamba}

Mamba \cite{mamba} is a novel deep sequence model architecture addressing the computational inefficiency of traditional Transformers on long sequences. It is based on selective state space models (SSMs), which improve upon previous SSMs by allowing the model parameters to be input functions. Here, some key concepts related to Mamba are explained. 

\subsubsection{Selective State Space Model}

Consider a structured SSM mapping one-dimensional sequence $x(t) \in \mathbb{R}^L$ to $ y(t) \in \mathbb{R}^L$ through a hidden state $h(t) \in \mathbb{R}^N$. With the evolution parameter $\vA \in \mathbb{R}^{N \times N}$ and the projection parameters $\vB \in \mathbb{R}^{N \times 1}$, $\vC \in \mathbb{R}^{1 \times N}$, such a model is formulated using linear ordinary differential equations
\begin{align}
\begin{split}
    h'(t) = &~ \vA h(t) + \vB x(t), \\
    y(t) = &~ \vC h(t).      
\end{split}
\label{eq:SSM}
\end{align}

As a continuous-time model, SSM is discretized with a Zero-Order Hold (ZOH) assumption to adapt to deep learning. Therefore, the continuous-time parameters $\vA, \vB$ are transformed to their discretized counterparts $\overline{\vA}, \overline{\vB}$ with a timescale parameter $\Delta$ according to

\begin{align}
\begin{split}
    \overline{\vA} = &~\exp (\Delta \vA),\\
    \overline{\vB} = &~ (\Delta \vA)^{-1}(\exp(\Delta \vA)-\vI) \cdot \Delta \vB.
\end{split}
\label{eq:dis-A-B}
\end{align}

Thus, (\ref{eq:SSM}) can be rewritten as
\begin{align}
\begin{split}
    h_t =&~ \overline{\vA} h_{t-1} + \overline{\vB} x_t, \\
    y_t = &~ \vC h_t.
\end{split}
\label{eq:dis-SSM}
\end{align}

To enhance computational efficiency and scalability, the iterative process in (\ref{eq:dis-SSM}) can be synthesized through a global convolution
\begin{align}
\begin{split}
    \overline{\vK} =&~ (\vC\overline{\vB},\vC\overline{\vA}\overline{\vB},\cdots, \overline{\vA}^{L-1}\overline{\vB}),\\
    \vy = &~ \vx * \overline{\vK},
\end{split}
\label{eq:conv-SSM}
\end{align}
where $L$ is the length of the input sequence $\vx$, $\overline{\vK} \in \mathbb{R}^L$ serves as the kernel of the SSM and $*$ represents the convolution operation.

Traditional SSM demonstrated linear time complexity, but their representativity of sequence context is inherently limited by time-invariant parameterization. To overcome the existing constraint, Selective SSM introduces a selective scan for interactions among sequential states with 
\begin{equation}
    \begin{split}
       \vB =& S_{\vB}(\vx), \\
       \vC =& S_{\vC}(\vx), \\
       \Delta =& \tau_\Delta( \Delta + S_\Delta(\vx)),
   \end{split}
   \label{eq:s6}
\end{equation}
before (\ref{eq:dis-A-B},\ref{eq:dis-SSM}), so that parameters $\vB \in \mathbb{R}^{B\times L \times N}$, $ \vC^{B\times L \times N}$ and $\Delta^{B\times L \times D}$ are dependent on the input sequence $\vx \in \mathbb{R}^{B\times L \times D}$, where $B$ represents the batch size, and $D$ represents number of channels. Normally, $S_B$ and $S_C$ are linear parameterized projections to dimension $N$, that is, $Linear_N(\cdot)$, while $S_\Delta(\vx) = Broadcast_D(Linear_1(\vx))$ and $\tau_\Delta = softplus$. The choice of $S_\Delta$ and $\tau_\Delta$ is caused by a relationship with RNNs gating mechanisms explained later.

\subsubsection{Selection Mechanism}

There is a well-established link between discretizing continuous-time systems and RNNs gating~\cite{tallec2018can}. One example of the selection mechanism for SSM is the traditional gating mechanism of RNNs. When $N=1, \vA = -1, \vB = 1, S_\Delta = Linear(\vx)$ and $\tau_\Delta = softplus$, then the selective SSM recurrence takes the form:
\begin{align}
    \begin{split}
        g_t =&~ \sigma(Linear(x(t))),\\
        h_t =&~ (1-g_t)h_{t-1} + g_t x_t,
    \end{split}
\label{SelectionMechanism}
\end{align}

\subsubsection{Scan}

The selection mechanism is devised to address the constraints of Linear Time Invariance (LTI) models. However, it reintroduces the computation issue in association with SSM. To enhance GPU utilization and efficiently materialize the state $h$ within the memory hierarchy, hardware-aware state expansion is enabled by selective scan. By incorporating kernel fusion and recomputation with parallel scan, the fused selective scan layer can effectively decrease the quantity of memory I/O operations, leading to a significant acceleration compared to conventional implementations.

\section{Methodology}

\begin{figure*}[!t]
\centering
\includegraphics[width=\linewidth]{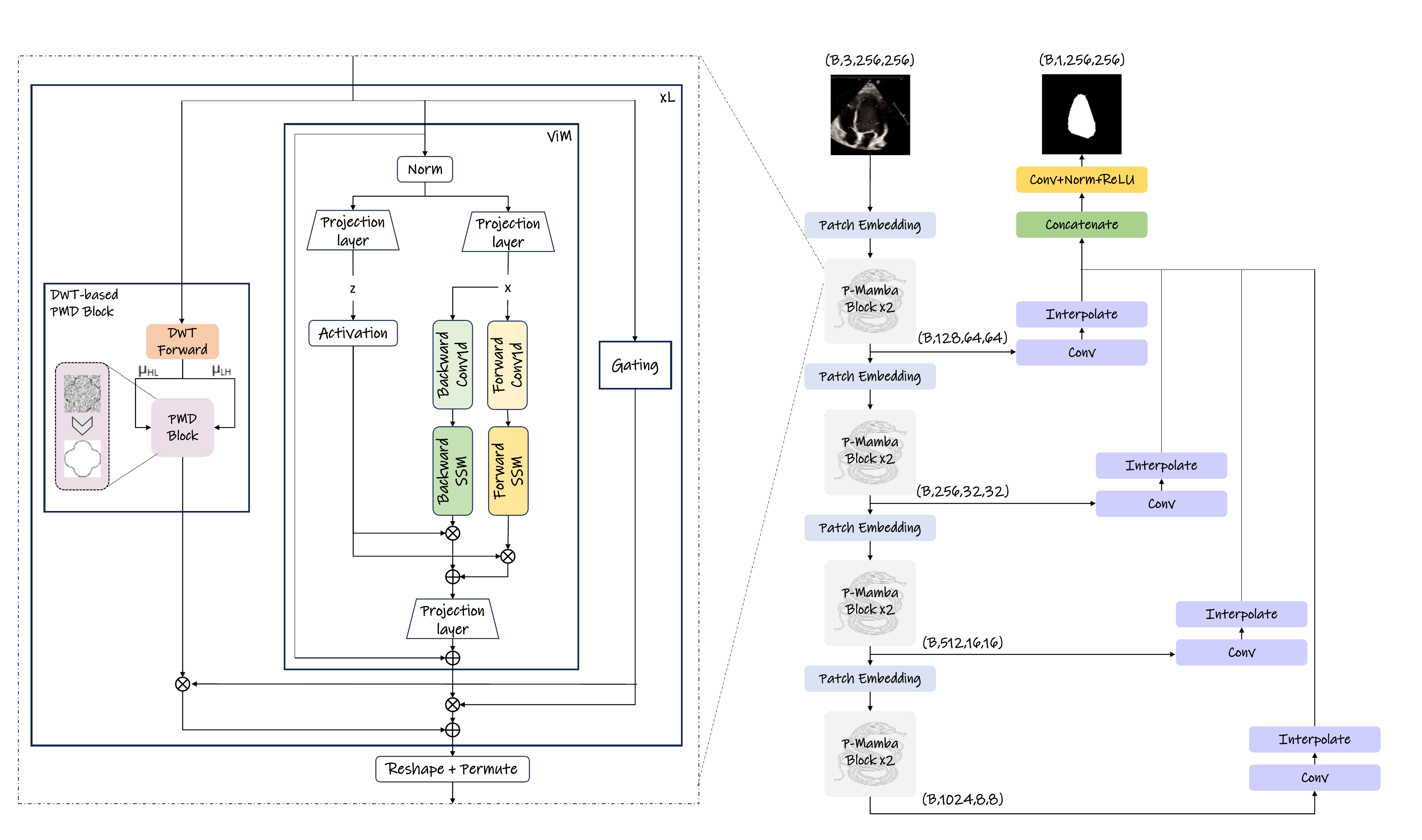}
\caption{The structure of our P-Mamba.
\label{fig:overview}}
\end{figure*}

We illustrate the overall architecture of P-Mamba in Fig.~\ref{fig:overview}. The P-Mamba network, designed for automatically segmenting the left ventricle in echocardiograms, leverages the Mixture of Experts framework. The DWT-based PMD block suppresses background noise interference while preserving target edges to capture more local features. The ViM block \cite{vim}, sometimes also mentioned as the Bidirectional Mamba block, is designed to guarantee the high efficiency of our model while encapsulating global dependencies. A gating mechanism inspired by the MoE structure dynamically selects appropriate pathways for feature processing, enhancing efficiency and segmentation accuracy. 

The P-Mamba encoder, incorporating multiple P-Mamba blocks, extracts hierarchical features at descending resolutions of 1/4, 1/8, 1/16, and 1/32 relative to the original input size while progressively increasing the channels at each level. In the following sections, we will introduce the detailed structure of P-Mamba.

\subsection{DWT-based PMD Block}

Initially used in image de-noising tasks, Perona-Malik Diffusion (PMD) can suppress noise disturbance while preserving boundary details. Considering that echocardiograms contain heavy background noise that may interfere with segmentation accuracy, we propose the DWT-based PMD Block to act on feature maps so that the background noise can be filtered. At the same time, some target boundary cues can be preserved.

Given an input feature map $u$, its PMD equation can be expressed as:
\begin{equation}\label{pmd eq}
\frac{\partial u}{\partial t}=div\left( g\left( |\nabla u| \right) \nabla u \right)
\end{equation}
where $g(|\nabla u|) = \frac{1}{1 + \left( \frac{|\nabla u|}{k} \right)^2}$ is the diffusion coefficient; $t$ is the diffusion step and can be regarded as the layer of the feature map; $k$ is a positive constant to control the degree of diffusion and is set to 1 by default in our experiments. Notably, \eqref{pmd eq} is an anisotropic diffusion equation. In the flat or smooth regions where the gradient magnitude is small ($|\nabla u|\rightarrow0$), the diffusion coefficient $g$ is large, meaning that the diffusion is strong and \eqref{pmd eq} acts as Gaussian smoothing to remove the noise interference. Somewhere near the target’s boundary, the gradient magnitude is large ($|\nabla u|\rightarrow1$), so the coefficient $g$ is near zero, meaning the diffusion is weak, and the boundary details can be preserved. Equation \eqref{pmd eq} can also be rewritten to the following form:
\begin{equation}\label{pmd1}
\begin{aligned}
\frac{\partial u}{\partial t}= & \frac{\partial}{\partial x}\left\{g\left(\sqrt{\left(\frac{\partial u_k}{\partial x}\right)^2+\left(\frac{\partial u_k}{\partial y}\right)^2}\right) \frac{\partial u_k}{\partial x}\right\} \\
& +\frac{\partial}{\partial y}\left\{g\left(\sqrt{\left(\frac{\partial u_k}{\partial x}\right)^2+\left(\frac{\partial u_k}{\partial y}\right)^2}\right) \frac{\partial u_k}{\partial y}\right\}
\end{aligned}
\end{equation}
where $\frac{\partial u}{\partial x}$ and $\frac{\partial u}{\partial y}$ represent the gradients of the feature map in horizontal and vertical directions. On the other hand, the Discrete Wavelet Transform (DWT) of an input feature map can be expressed as:
\begin{equation}\label{dwt}
u_{i}=DWT(u), i \in \left\{u_{LL},u_{LH},u_{HL},u_{HH}\right\}
\end{equation}
where $u_{LL}$ is the low-frequency part of the feature map, while $u_{LH}$, $u_{HL}$ and $u_{HH}$ are the high-frequency parts in horizontal, vertical and diagonal directions of the feature map which mainly contain the boundary details. By approximating the derivative terms $\frac{\partial u}{\partial x}$ with $u_{LH}$ and $\frac{\partial u}{\partial y}$ with $u_{HL}$ and setting the diffusion step size $\delta t$ to one, we can transform \eqref{pmd1} to the discrete format:

\begin{equation}
\begin{aligned}
u_k=  & u_{k-1}+\left[g\left(\sqrt{u_{L H}^2+u_{H L}^2}\right) \cdot u_{L H}\right]_{LH}\\
 & +\left[g\left(\sqrt{u_{L H}^2+u_{H L}^2}\right) \cdot u_{H L}\right]_{HL}
 \end{aligned}
\end{equation}

After enhancing the feature map by PMD, we feed the diffusion output into a basic ResNet block. Piling multiple DWT-based PMD blocks in all layers of the
encoder branch, our P-Mamba can suppress background noise disturbance while preserving the target boundary features.

\subsection{Vision Mamba Block}

We mainly adopt the ViM layer to improve our model's computing and memory efficiency. Also, it helps capture global dependencies complementary to the local shape cues extracted by our DWT-based PMD Block. 

Before processing into the P-Mamba block, a 2-D input $\in \mathbb{R}^{H \times W\times C}$ is transformed into flattened patches $P_N$ with dimensions $M \times (N^2 \cdot C)$. Here $(H, W)$ is the size of the original input, $C$ stands for the number of channels, and $N$ and $M$ denote the size and total count of segmented patches, respectively. Then, the $P_N$ is linearly projected to vectors of dimension D and add position embedding $E_{\text{pos}} \in \mathbb{R}^{M \times D}$. This process can be described as follows:

\begin{equation}
X_0 = [x^1W; x^2W; \dots; x^MW] + E_{\text{pos}}
\end{equation}
where \(x^M\) is the  \(M^{th}\) patch of \(P_N\), \(W \in \mathbb{R}^{(N^2 \cdot C) \times D}\) is the learnable projection matrix. The token sequence from the patch embedding layer,  \(X_{pe}\), is processed by the layer ViM to obtain the output \(X_{vim}\), which is expressed by \eqref{vim_eq}. 

\begin{equation}\label{vim_eq}
X_{vim} = Vim(X_{pe}) + X_{pe}
\end{equation}

Subsequently, the outputs from the DWT-based PMD module and the ViM module are iteratively combined using the gating mechanism across several layers within the p-Mamba block. The final output is then reshaped and permuted to serve as the input for the next stage, where the process is repeated to obtain the next stage's output.

\subsection{Decoder}

The decoder consists of multiple interpolation layers interspersed with Conv2d operations, progressively upsampling the encoded features to generate high-resolution segmentation maps. Specifically, the decoder processes the output feature map from the different encoder stages through several Conv2d layers, followed by interpolation operations that align the feature maps to a common resolution. These feature maps are then concatenated and passed through a series of Conv2d, batch normalization, and ReLU layers to produce the final segmentation map.

\section{Experiment}

\subsection{Datasets}

\subsubsection{Pediatric Dataset}

The dataset comprises echocardiographic evaluations from patients at Lucile Packard Children's Hospital Stanford from 2014 to 2021, authorized by the Stanford University Institutional Review Board. The dataset contains 4467 echocardiograms collected from 1958 patients, 43$\%$ female, aged between 0 and 18 years (mean ± SD: 10 ± 5.4 years). The patients were classified into two groups based on their echocardiographic results: those with structurally normal hearts and average ejection fraction (EF) and those with structurally normal hearts but systolic dysfunction (including dilated cardiomyopathy, chemotherapy-induced systolic dysfunction) without congenital heart disease \cite{b20}. 

After additional processing, the dataset was employed to obtain apical four-chamber (A4C) and parasternal short-axis (PSAX) video clips, totaling 7643 video clips and 17600 annotated pictures. The video clips were partitioned into training (80$\%$, n=6114), testing (10$\%$, n=765), and validation (10$\%$, n=764) sets for machine learning purposes. In addition, 86$\%$ of the trials had an ejection fraction (EF) equal to or greater than 55$\%$.

\subsubsection{EchoNet-Dynamic Dataset}

To ensure the model's effectiveness on general echocardiogram datasets, not just pediatric ones, we also conducted experiments on the EchoNet-Dynamic dataset \cite{dynamic}. EchoNet-Dynamic, obtained from Stanford University Hospital, is the largest publicly available dataset of apical four-chamber (A4C) cardiac echocardiograms. It includes 10030 echocardiogram clips and 9989 video samples that remained after data cleansing. Each video was examined solely for the end-diastolic and end-systolic frames. As a result, 96 images were excluded due to poor-quality ground truth, leaving 14846 out of 19882 images for the training set. The remaining images were divided into validation (2563 images) and testing (2473 images) sets. End-diastolic and end-systolic frames from the same individuals were grouped for the experiments.

\subsection{Implementation Details}

The computational setup consists of a single Tesla V100-32GB GPU, a 12-core CPU, and 61GB of RAM. The system operates on an Ubuntu 18 environment with CUDA 11.0 and Pytorch 1.13 software.

The network was trained for 50 epochs, beginning with an initial learning rate 1e-4. Batch sizes of 24 were selected for training to obtain a compromise between computational efficiency and model accuracy. The model's performance was assessed every five epochs, and early halting with a patience parameter of 10 was implemented to prevent overfitting. The network architecture was organized with layers configured in depth [2, 2, 2, 2].

\section{Results and Discussion}

Table~\ref{tab1} offers a quantitative comparison of our P-Mamba with various state-of-the-art methods on the pediatric LV 2D segmentation task and includes experiments on the general ultrasound dataset EchoNet-Dynamic. The comparison includes CNN-based methods like U-Net \cite{ref_gupta2023} with FCN and PSPNet backbones, as well as ViT-based approaches such as PVT \cite{ref_wang2021}, Flatten Transformer \cite{ref_han2023}, and MaxVit \cite{ref_tu2022}. Moreover, we also compared Uniformer \cite{ref_li2023} and SpectFormer \cite{ref_patro2023}, both newly introduced segmentation models from the last year, showcasing advanced capabilities in image segmentation. U-Mamba \cite{umamba} represents a straightforward plan Mamba model characterized by a U-shaped structure designed specifically for effective segmentation tasks. TransUnet \cite{transunet} and Segformer \cite{segformer} are recognized as classic segmentation models, and they are widely used in the field due to their robust performance. Additionally, CMUNet \cite{cmu} and NHBS \cite{nhbs} are specialized models tailored for ultrasound image segmentation, addressing the unique challenges of medical imaging.

As a result, Table~\ref{tab1} demonstrates the superiority of our P-Mamba model. It achieves the highest average Dice Similarity Coefficient (DSC) on the PSAX and A4C pediatric datasets, with values of 0.9221 and 0.9056, respectively. Furthermore, our model also excels on the EchoNet-Dynamic dataset, with an impressive DSC of 0.9314. These results highlight the effectiveness and robustness of our approach across different datasets, outperforming the listed state-of-the-art methods.

\renewcommand{\arraystretch}{1}  
\begin{table*}[htbp]
\small
\centering
\caption{Comparison with the state-of-the-art methods.}\label{tab1}
\setlength{\tabcolsep}{1mm}
\begin{tabular}{c | c | c | c | c | c | c | c | c | c}
    \toprule
    \multirow{2}{*}{Methods} & \multicolumn{3}{c|}{Pediatric PSAX} & \multicolumn{3}{c|}{Pediatric A4C} & \multicolumn{3}{c}{EchoNet-Dynamic}\\
    & Precision & Recall & Dice & Precision & Recall & Dice & Precision & Recall & Dice\\
    \midrule
    UNet (FCN) \cite{ref_gupta2023}& 0.8492 & 0.8761 & 0.8624 & 0.8279 & 0.8345 & 0.8312&0.8837 & 0.8773 & 0.8805\\
    UNet (PSPNet) \cite{ref_gupta2023}& 0.8651 & 0.8700 & 0.8675 & 0.8204 & 0.8607 & 0.8401&0.8810 &0.8899 &0.8854\\
    PVT \cite{ref_wang2021} &0.8731 &  0.8371 &   0.8547 & 0.8670 & 0.8366 & 0.8515&  0.9165& 0.9037&0.9101 \\
    UniFormer  \cite{ref_li2023}& 0.9100 & 0.9134 & 0.9073 & 0.8969 & 0.8915 & 0.8918&0.896 & 0.9157 &  0.9058  \\
    SpectFormer \cite{ref_patro2023}& 0.9127 & 0.9063 & 0.9161 & 0.9076 & 0.9035 & 0.9010& 0.9195 &  0.9234&0.9214 \\
    Flatten Transformer \cite{ref_han2023} & 0.9215 & 0.9122 & 0.9168 & 0.9130 & 0.8832 & 0.8978&0.9218  & 0.9277&0.9247 \\
    MaxVit \cite{ref_tu2022} & 0.9024 & 0.9249 & 0.9135 & 0.8912 & 0.9066 & 0.8988&0.9246 &0.9280  & 0.9263\\
    U-Mamba  \cite{umamba} & 0.9275  & 0.9017 & 0.9144 &  0.9022 & 0.8857 & 0.8939  &0.9331 &0.9156 &0.9243 \\
    TransUnet \cite{transunet} &0.8555 & 0.8059 & 0.8299 &0.8273  &  0.8390 &  0.8331&0.9264 & 0.8991& 0.9084\\
    Segformer \cite{segformer} & 0.8646 &0.8345  & 0.8493 & 0.8481 &0.8089 & 0.8280 &  0.8917 & 0.8886  & 0.8902 \\
    CMUNet  \cite{cmu} &  0.9197 &  0.9126 & 0.9161 &  0.9073 & 0.8959 & 0.9016 &  0.9303 & 0.9266 &  0.9285  \\
    NHBS  \cite{nhbs} & 0.8734 & 0.8901 & 0.8817 & 0.8864 & 0.8723 & 0.8793 &  0.9224 & 0.9159 &  0.9192 \\   
    Ours & \textbf{0.9316} & 0.9128 & \textbf{0.9221} & 0.9045 & \textbf{0.9067} & \textbf{0.9056}& 0.9186&\textbf{0.9446} &\textbf{0.9314} \\
    \bottomrule
\end{tabular}
\end{table*}

\begin{figure}[htbp]
\centering
\includegraphics[width=1\linewidth,height=0.84\textwidth]{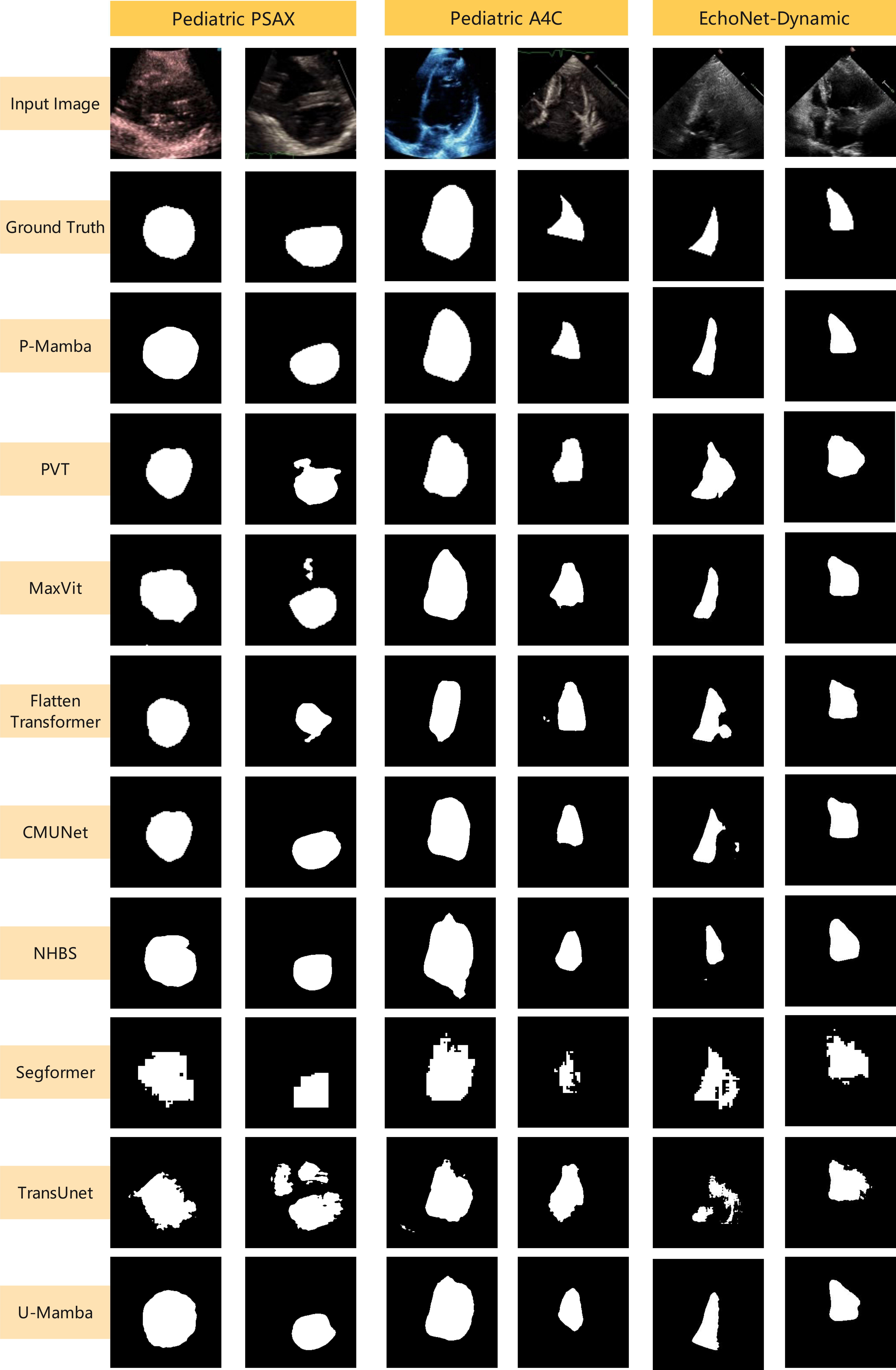}
\caption{The visual comparison of different methods.}
\label{fig4}
\end{figure}

\begin{figure}[htbp]
\centering
\includegraphics[width=1\linewidth]{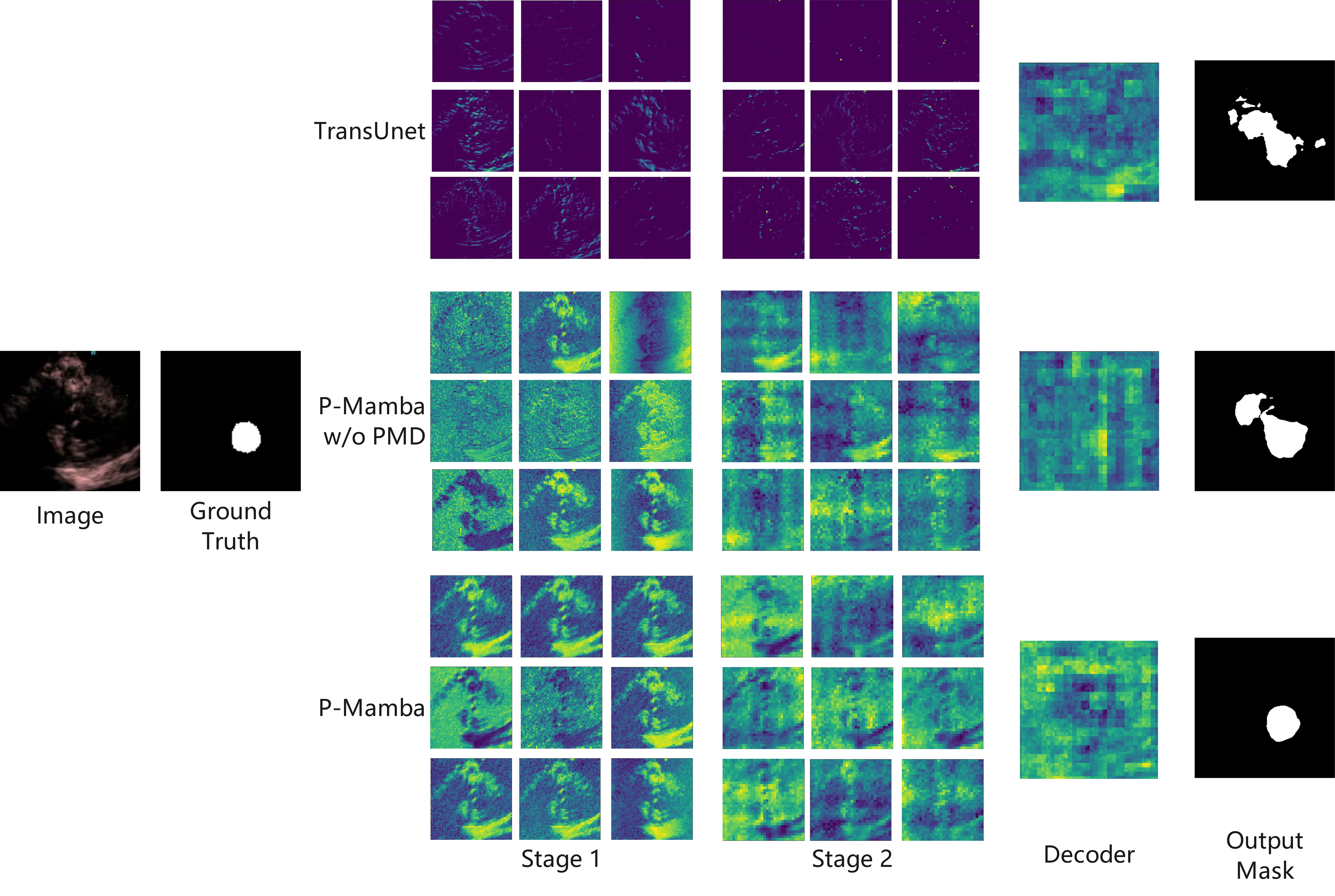}
\caption{The Visualization of the features of TransUnet, P-Mamba w/o PMD, and P-Mamba.}
\label{featuremap}
\end{figure}

\subsection{Ablation Studies}

We first ablate the DWT-based PMD Block in Table~\ref{tab2} across various configurations: 'Ours w/o PMD' means removing the DWT-based PMD part; 'Ours w/ Sobel' means replacing the DWT-based PMD part with a Sobel operator, which is only for edge preservation. Our DWT-based PMD block achieves the best performance on those three datasets. The DWT-based PMD part encompasses the Sobel operator due to its noise suppression and edge preservation capability, while the Sobel operator cannot finely preserve edges during noise removal.

In addition, Table~\ref{tab3} studies the effect of the Vision Mamba block. We replaced the ViM block with ViT structures of quadratic (PVT) and linear complexity (Flatten ViT, MaxViT). The results, summarized across the PSAX, A4C, and EchoNet-Dynamic datasets, indicate that Vision Mamba consistently achieves a higher Dice coefficient. Our findings indicate that Vision Mamba can be more accurate than other models with linear or quadratic complexity.

To further validate the effectiveness of the mixture of expert's gating network convergence, we compared it with the results of simply adding the outputs of the DWT-based PMD Block and the ViM module. Table~\ref{converge} shows that the gating structure improves the Dice coefficient across three datasets, demonstrating that the gating structure implemented with a linear layer significantly improves precision. 

\begin{table*}[htbp]
\small
\centering
\caption{Ablation study on the DWT-based PMD Block design.}\label{tab2}
\centering
\label{ab1}
\setlength{\tabcolsep}{1mm}
\begin{tabular}{c | c | c | c | c | c | c | c | c | c}
    \toprule
     \multirow{2}{*}{Methods} & \multicolumn{3}{c|}{PSAX Dataset} & \multicolumn{3}{c|}{A4C Dataset} & \multicolumn{3}{c}{EchoNet-Dynamic}\\
    & Precision & Recall & Dice & Precision & Recall & Dice& Precision & Recall & Dice \\
    \midrule
    Ours w/o PMD & 0.8902 & 0.9098 & 0.8999 & 0.8795 & 0.8811 & 0.8803&0.9215 & 0.8931& 0.9071\\
    Ours w/ Soble & 0.9312 & 0.9013 & 0.9160 & 0.9073 & 0.8983 & 0.9028 & 0.9237& 0.9166& 0.9201\\
    Ours & \textbf{0.9316} & 0.9128 & \textbf{0.9221} & 0.9045 & \textbf{0.9067} & \textbf{0.9056} &0.9186&\textbf{0.9446} &\textbf{0.9314} \\
    \bottomrule
\end{tabular}
\end{table*}

\begin{table*}[htbp]
\small
\centering
\caption{Ablation study on the ViM Block design.}\label{tab3}
\centering
\label{ab2}
\setlength{\tabcolsep}{1mm}
		\begin{tabular}{c | c | c | c | c | c | c | c | c | c}
    \toprule
        \multirow{2}{*}{Methods} & \multicolumn{3}{c|}{PSAX Dataset} & \multicolumn{3}{c|}{A4C Dataset} & \multicolumn{3}{c}{EchoNet-Dynamic}\\
        &Precision & Recall & Dice &Precision & Recall & Dice& Precision & Recall & Dice\\
        \midrule
Mamba $\rightarrow$ PVT & 0.9029 & 0.9183 & 0.9105 & 0.9045 & 0.8714 & 0.8876 & 0.9147 &0.8907 &0.9025 \\ 
Mamba $\rightarrow$ Flatten Transformer & 0.9137 & 0.9157 & 0.9147 & 0.9228 & 0.8752 & 0.8984 & 0.9198&0.8974 &  0.9085\\ 
Mamba $\rightarrow$ MaxVit & 0.9253 & 0.9087 & 0.9169 & 0.9062 & 0.9013 & 0.9038&  0.9211 &0.9004 & 0.9106 \\ 
Ours            & \textbf{0.9316} & 0.9128 & \textbf{0.9221} & 0.9045 & \textbf{0.9067} & \textbf{0.9056}&0.9186&\textbf{0.9446} &\textbf{0.9314} \\ 
        \bottomrule
    \end{tabular}\label{ab4}
\end{table*}

\begin{table*}[htbp]
\small
\centering
\caption{Ablation study on the converge design.}\label{converge}
\centering
\label{ab2}
\setlength{\tabcolsep}{1mm}
		\begin{tabular}{c | c | c | c | c | c | c | c | c | c}
    \toprule
        \multirow{2}{*}{Methods} & \multicolumn{3}{c|}{PSAX Dataset} & \multicolumn{3}{c|}{A4C Dataset} & \multicolumn{3}{c}{EchoNet-Dynamic}\\
        &Precision & Recall & Dice &Precision & Recall & Dice& Precision & Recall & Dice\\
        \midrule
Adding & 0.9203 & 0.9200  & 0.9201 & 0.9090 &0.8826 & 0.8956 & 0.9320  & 0.9270& 0.9295 \\ 
Ours (Gating) & \textbf{0.9316} & 0.9128 & \textbf{0.9221} & 0.9045 & \textbf{0.9067} & \textbf{0.9056}&0.9186&\textbf{0.9446} &\textbf{0.9314} \\ 
        \bottomrule
    \end{tabular}\label{ab4}
\end{table*}

\subsection{Model Efficiency Comparison}

Table~\ref{eff} presents the model efficiency comparison results of P-Mamba with various state-of-the-art methods, considering parameters, inference speed (ms), GPU memory usage (GB), and GFLOPs. Our P-Mamba significantly outperforms other methods across all metrics. Specifically, P-Mamba has the lowest parameter count (52.77M), fastest inference speed (9.18 ms), least GPU memory usage (4.95 GB), and lowest GFLOPs (24.66). The attention-free design of our Mamba block greatly enhances model efficiency, even compared to models with linear complexity. Additionally, our DWT-based PMD block does not add excessive parameters compared to the pure mamba net like U-Mamba, ensuring high efficiency. 

\begin{table}[htbp]
\caption{Model efficiency comparison regarding parameter number (M), inference speed (ms), GPU memory (GB), and GFLOPs.}\label{eff}
\centering
\setlength{\tabcolsep}{0.5mm}
\begin{tabular}{c  c  c  c  c}
    \toprule
    Methods & \#Params & ms/Inf & GPU memory & GFLOPs\\
    \midrule
    PVT \cite{ref_wang2021} & 299.81 & 40.39 & 15.39 & 181.88\\
    Flatten Transformer \cite{ref_han2023} & 333.43 & 86.12 & 15.55 & 184.08\\
    MaxVit \cite{ref_tu2022} & 480.79 & 48.49& 20.08 & 259.48\\
    U-Mamba  \cite{umamba} & 74.00 & 18.64 & 8.38 & 37.14\\
    TransUnet \cite{transunet}  &77.08 & 23.82& 15.67  &36.69\\
    Segformer \cite{segformer} & 77.20& 24.07 &19.48 & 38.90\\
    CMUNet  \cite{cmu}  &149.93 &30.22 &  23.54&182.68\\
    NHBS  \cite{nhbs}  & 102.20  & 21.64 & 19.35 & 119.49\\
    Ours & \textbf{52.77} & \textbf{9.18} & \textbf{4.95} & \textbf{24.66}\\
    \bottomrule
\end{tabular}
\end{table}

\subsection{Qualitative Comparison}

The qualitative results in Fig.~\ref{fig4} illustrate the performance of various models on pediatric PSAX, pediatric A4C, and EchoNet-Dynamic image segmentation tasks. Classic segmentation models like Segformer and TransUnet display poor segmentation results, indicating their unsuitability for echocardiogram data modalities. Due to noise interference, PVT, Flatten Transformer, MaxVit, and U-Mamba struggle to delineate heart structures correctly. Specialized ultrasound segmentation models such as NHBS and CMUNet also fall short in accurately segmenting the A4C views. In contrast, our P-Mamba is the least affected by noise interference and enjoys the best segmentation effect, thanks to our PMD design. 

We visualize the learned features of TransUnet, P-Mamba without the DWT-based PMD Block, and P-Mamba to better understand the impact of the proposed method. The feature map outputs of encoder stages 1 and 2 are shown in Fig.~\ref{featuremap}, with nine feature map channels randomly sampled for each method. With more distinct feature contours, local information is better preserved in P-Mamba than in other methods. Additionally, we visualize the feature map output of the final decoder stage in Fig.~\ref{featuremap} to support our analysis. We observe that P-Mamba features are more concentrated and exhibit stronger discriminative power.

\section{Conclusion}

We present P-Mamba, a model tailored for efficient left ventricular segmentation in pediatric echocardiography, overcoming the background noise interference challenge. Specifically, the DWT-based Perona-Malik Diffusion Block is a mathematically explainable module focusing on local feature extraction. It can gradually suppress background noise disturbance while preserving the boundary details of the left ventricle. The vision Mamba block, on the other hand, can explore global dependencies. We integrate the two parts by borrowing the Mixture of Experts (MOE) ideas, and our P-Mamba successfully achieves an exceptional balance between segmentation accuracy and efficiency.

\bibliographystyle{IEEEtran}
\bibliography{main}

\end{document}